\documentclass[runningheads]{llncs}
\usepackage{esvect}
\usepackage[T1]{fontenc}
\usepackage{graphicx}
\usepackage{kotex}
\usepackage{times}
\usepackage{latexsym}
\usepackage{courier}
\usepackage{multirow}
\usepackage{makecell}
\usepackage{amsfonts}
\usepackage{amsmath}
\usepackage{subcaption}
\usepackage{tipa}
\usepackage{fixltx2e}
\usepackage{pifont} 
\usepackage{tablefootnote}
\usepackage{bbold}
\usepackage{booktabs}
\usepackage{float}
\usepackage{wrapfig}
\usepackage{caption}  

\begin{document}

\title{Handling Korean Out-of-Vocabulary Words with Phoneme Representation Learning}

\author{Nayeon Kim\textsuperscript{1†}\orcidID{0000-0002-7971-1627} \and
Eojin Jeon\textsuperscript{1†}\orcidID{0000-0002-7503-987X} \and \\
Jun-Hyung Park\inst{2}\orcidID{0000-0002-7900-3743} \and
SangKeun Lee\inst{1*}\orcidID{0000-0002-6249-8217}}

\authorrunning{Kim et al.}

\institute{Korea University, Seoul, South Korea\\
\email{\{lilian1208, skdlcm456, yalphy\}@korea.ac.kr} \and
Hankuk University of Foreign Studies, Seoul, South Korea\\
\email{jhp@hufs.ac.kr}}
\maketitle        

\def\thefootnote{†}\footnotetext{Equal contribution.}\def\thefootnote{\arabic{footnote}}
\def\thefootnote{*}\footnotetext{Corresponding author.}\def\thefootnote{\arabic{footnote}}
\begin{abstract}
In this study, we introduce KOPL, a novel framework for handling \textbf{K}orean \textbf{O}OV words with \textbf{P}honeme representation \textbf{L}earning. Our work is based on the linguistic property of Korean as a phonemic script, the high correlation between phonemes and letters. KOPL incorporates phoneme and word representations for Korean OOV words, facilitating Korean OOV word representations to capture both text and phoneme information of words. We empirically demonstrate that KOPL significantly improves the performance on Korean Natural Language Processing (NLP) tasks, while being readily integrated into existing static and contextual Korean embedding models in a plug-and-play manner. Notably, we show that KOPL outperforms the state-of-the-art model by an average of 1.9\%. Our code is available at \url{https://github.com/jej127/KOPL.git}.

\end{abstract}
\section{Introduction}

Handling Out-of-Vocabulary words, which are the words unseen during training, remains a longstanding challenge, as they can degrade the performance of downstream tasks~\cite{Jin_AAAI2020,Liang_IJCAI2018,Sun_Corr2020}. Various approaches have dealt with OOV words based on pre-training subword-level representation from scratch~\cite{fastText_bojanowski17,kim-etal-2022-break,park-etal-2018-subword} or on encouraging learned representations for OOV words to be close to well-trained word embeddings~\cite{LOVE_chen22,GRM-2023-graph,MIMICK_pinter17,KVQFH_sasaki19,BOS_zhao18}. However, since the formation of OOV words could be language-dependent, existing approaches focusing on English OOV words can struggle to address OOV words in other languages including Korean. 

In this paper, we focus on handling Korean OOV words by considering the linguistic property of Korean, the high correlation between phonemes and letters. Our work is aligned with the recent studies that have considered Korean-specific linguistic knowledge such as the writing system~\cite{kim-etal-2022-break,kim-etal-2024-kombo,park-etal-2018-subword} or syntactic features~\cite{lee-etal-2024-length,Park_AACL2020,park2021klue,seo-etal-2023-chef}, to improve the performance of various Korean NLP tasks. Since the Korean letter \textit{Hangeul} as a phonemic script~\cite{sampson1985writing} has a systematic internal structure correlated with the feature of phonemes, phonemes can contain useful information to understand Korean. Regarding Korean OOV words, this property of \textit{Hangeul} makes us pay attention to the theory of psycholinguistics, the interactions between phonemes and letters in forming the meaning of words~\cite{psychology_carroll1986}.

\setlength{\intextsep}{10pt}  
\setlength{\columnsep}{15pt} 

\begin{wrapfigure}{r}{0.5\textwidth}  
  \includegraphics[width=0.5\textwidth]{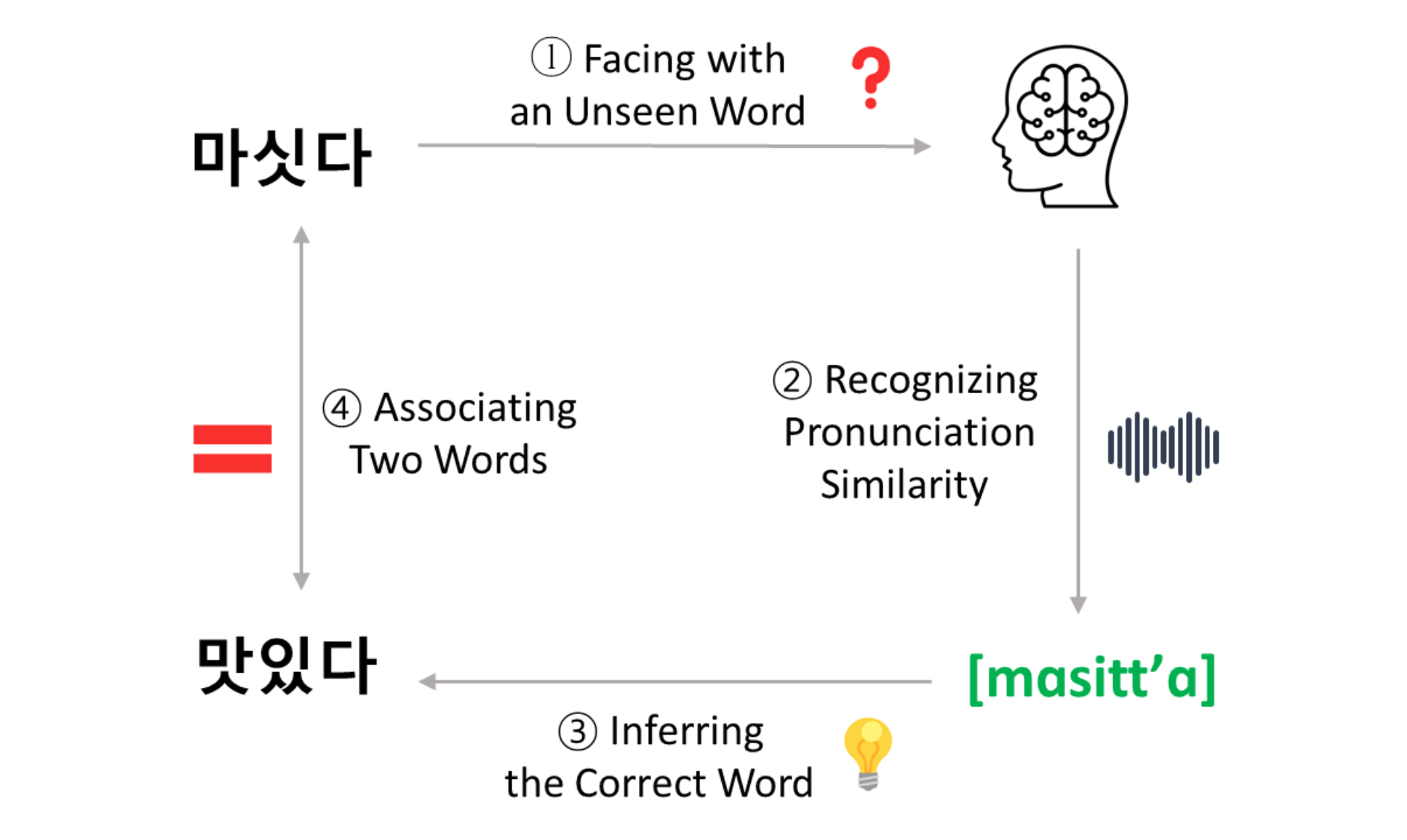} 
 \caption{Illustration of humans' forming the word meaning. When encountering the unseen word ``*마싯다[\textipa{mAsitt'A}]\textsubscript{*dilisious}'', humans can recall its phoneme information ``[\textipa{mAsitt'A}]'' to infer the correct word ``맛있다[\textipa{mAsitt'A}]\textsubscript{delicious}''.}
  \label{motive}
\end{wrapfigure} 

For instance, Figure \ref{motive} shows one possible case where phoneme information plays a crucial role in understanding the meaning of words. When faced with ``*마싯다[\textipa{mAsitt'A}]\textsubscript{*dilisious}''\footnote{Asterisk(*) denotes ungrammatical words.}, a misspelling of ``맛있다[\textipa{mAsitt'A}]\textsubscript{delicious}'', humans can recognize its spelling and subsequently recall its pronunciation. They can then infer ``맛있다\textsubscript{delicious}'', the similarly pronounced word, as the correct word. 

Based on these observations, we introduce KOPL, a novel framework for handling \textbf{K}orean \textbf{O}OV words with \textbf{P}honeme representation \textbf{L}earning. KOPL learns representations of phonemes and words, and combines them to connect both types of representations with the word meaning. In addition, we propose two key strategies: phoneme-aware multimodal learning and cross-modal ensemble techniques.
Experimental results demonstrate that KOPL outperforms the previous state-of-the-art model by an average of 1.9\% across five Korean NLP downstream tasks featuring real-world OOV examples. KOPL can be seamlessly integrated into existing static and contextual Korean embedding models in a plug-and-play fashion.

To the best of our knowledge, KOPL is the first work to leverage phoneme information to improve Korean OOV word embeddings. Although there have been works to utilize phoneme information to make NLP models robust to input perturbations, these methods require either pre-training the dedicated transformers on the large-scale corpora~\cite{rocbert_acl2022,phonemeBERT}, or the separate detection and correction of misspellings~\cite{Li_EMNLP2022,Liang_ACL2023,Zhang_ACL2021}. Other works use phoneme representations by themselves or combining them with speech representations to address the sound sequence alignment~\cite{sofroniev-coltekin-2018-phonetic} or forced alignment problem~\cite{zhu2024taste}. Unlike these works, we focus on how to jointly utilize phoneme and text representations to address the OOV problem in Korean.

In summary, we introduce KOPL, a novel framework to enhance Korean OOV word representations based on the linguistic property of Korean, the high correlation between phonemes and letters. We propose novel phoneme-aware multimodal learning and cross-modal ensemble strategies to effectively combine textual and auditory modalities in downstream tasks. Moreover, KOPL outperforms previous OOV methods on various Korean NLP tasks and can be employed in a plug-and-play manner to improve existing Korean embedding models.

\section{Methodology}
\label{sec:method}

\begin{figure*}[t!]
\centering
  \includegraphics[width=\textwidth]{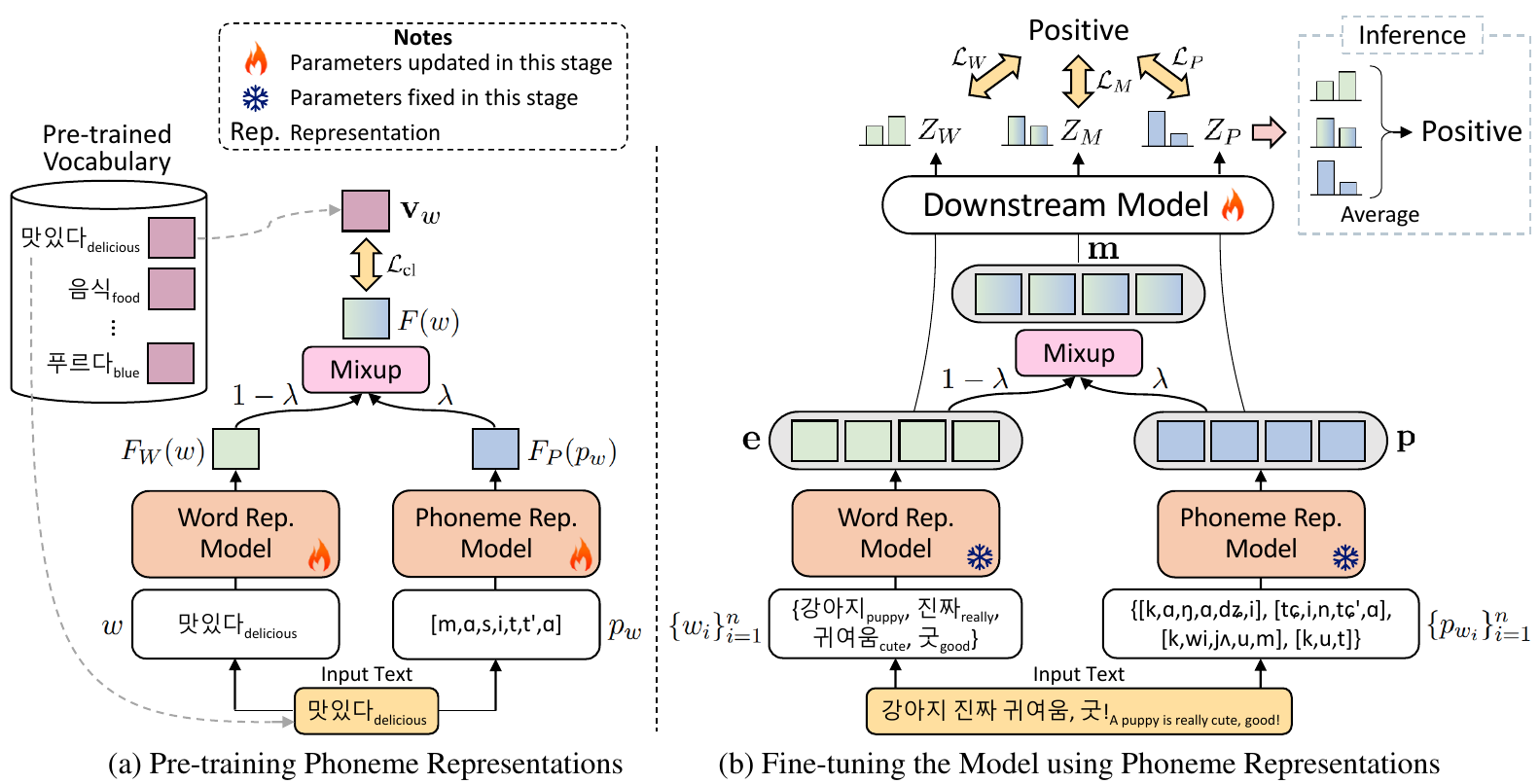}
    \caption{Illustration of KOPL. The meanings of inputs ``맛있다'' and ``강아지 진짜 귀여움, 굿!'' are ``delicious'' and ``A puppy is really cute, good!'', respectively. (a) In the pre-training stage, KOPL jointly learns phoneme and word representations using linear interpolation. (b) In the fine-tuning stage, KOPL uses phoneme, word, and mixed representations as separate inputs to the downstream model. During inference for downstream tasks, we average the prediction scores obtained from each type of representation to make the final prediction.}
    \label{fig:pipeline}
\end{figure*}

This section details our proposed KOPL. The core idea of KOPL is to learn phoneme representations for words that may not be present in a vocabulary (Section \ref{subsec:phonemevec}), and then to use these representations to fine-tune the models (Section \ref{subsec:utilphoneme}). An overview of KOPL is illustrated in Figure \ref{fig:pipeline}.

\subsection{Pre-training Phoneme Representations}
\label{subsec:phonemevec}
In this section, we describe how to pre-train phoneme representations. First, we consider a vocabulary $\mathcal{V}$ with a fixed size, and the pre-trained word embedding $\mathbf{v}_{w} \in \mathbb{R}^{d}$ for each word $w \in \mathcal{V}$. We aim to learn phoneme and word representations that not only capture the unique characteristics of phonemes and words but also contain complementary information present in both. To this end, we jointly train two neural networks $F_{P}$ and $F_{W}$, which output phoneme and word representations as separate $d$-dimensional vectors, respectively. We use a self-attention module with positional embeddings~\cite{Vaswani_NIPS17} for both neural networks $F_{P}$ and $F_{W}$. The input $p_{w}$ for $F_{P}$ is a list of phonemes corresponding to the word $w$,\footnote{We use the hangul\_to\_ipa program (\url{https://github.com/stannam/hangul\_to\_ipa}) to obtain phoneme sequence.} which is constructed by International Phonetic Alphabet (IPA) symbols~\cite{IPA_1999}. Furthermore, $F_{W}$ takes a mixed sequence of subcharacters and morphemes constituting $w$, which is the same as the input method for the Korean version of LOVE~\cite{LOVE_chen22} detailed in Table \ref{tab:models}.

To ensure that the two types of representation contain complementary information in phonemes and words, we seek to fuse them during pre-training. To this end, we interpolate the phoneme representation $F_{P}(p_{w})$ with the word representation $F_{W}(w)$:
\begin{equation}
\label{eq1}
    F(w)=\lambda F_{P}(p_{w}) + (1-\lambda) F_{W}(w),
\end{equation}
where $0 \le \lambda \le 1$ is a hyperparameter to derive the mixup ratio of each representation. The advantage of the interpolation is to balance the weights between phoneme and word representations, enabling their effective combination. While some words have similar meaning and pronunciation but different spellings, others have similar pronunciation\footnote{In this work, the term 'similar pronunciation' can be defined quantitatively, such as through the edit distance of IPA sequences or acoustic similarity of sound waves.} but different meanings. For instance, ``맛있다\textsubscript{delicious}'' and ``*마싯다\textsubscript{*dilisious}'', both pronounced as [\textit{m\textipa{A}sitt'\textipa{A}}], illustrate the former case. ``찾다\textsubscript{find}'' and ``찼다\textsubscript{kicked}'', both pronounced as [\textipa{t\textctc\super{h}Att'A}], exemplify the latter case. Although ``찾다" and ``찼다" have distinct meanings, their phoneme representations can be similar due to their similar pronunciation. Since the phoneme representations in such cases may struggle to distinguish the semantic differences between words, we use interpolation to complement semantic distinctions and learn richer representations. 

After fusing both types of representations, $F_{P}$ and $F_{W}$ are trained to predict the pre-trained word embedding $\mathbf{v}_{w}$ by using the vector output $F(w)$. Following the prior works \cite{LOVE_chen22,GRM-2023-graph}, we use a contrastive loss defined as follows, to train $F_{P}$ and $F_{W}$:
\begin{equation}
\label{eq2}
    \mathcal{L}_{\text{cl}}=-\frac{1}{|\mathcal{V}|}
    \sum_{w \in \mathcal{V}} \log{\frac{e^{s(F(w),\mathbf{v}_{w})/\tau}}{e^{s(F(w),\mathbf{v}_{w})/\tau} \!+ \!\sum e^{s(F(w),\mathbf{v}^{-})/\tau}}},
\end{equation}

where $s(\cdot,\cdot)$ denotes a scoring function such as the dot product and $\tau$ is a temperature parameter. $(F(w),\mathbf{v}^{-})$ stands for a negative pair, where $\mathbf{v}^{-}$ is a word embedding corresponding to other words in the same mini-batch. During this process, the modules $F_{P}$ and $F_{W}$ learn how to combine their output representations to predict the pre-trained word embedding. After training $F_{P}$ and $F_{W}$, we can obtain both phoneme and word representations for an arbitrary word $w$.

\subsection{Fine-tuning Downstream Models}
In this section, we describe two strategies for effectively leveraging phoneme representations in downstream tasks.
\label{subsec:utilphoneme}
\subsubsection{Phoneme-aware Multimodal Learning}
\label{subsubsec:multi_task}

 We consider a downstream task where the goal is to classify a sequence of words $\{w_{i}\}_{i=1}^{n}\!\in \mathcal{X}$ as a label $y \in \mathcal{Y}$, given the modules $F_{P}$ and $F_{W}$ that infer phoneme and word representations, respectively. The basic idea is to train the downstream model by feeding different types of representations as separate inputs into it, enabling the model to learn modality-specific features from both phonemes and texts. To be specific, we use the modules $F_{P}$ and $F_{W}$ to acquire a sequence of phoneme representations $\mathbf{p} = \{F_{P}(p_{w_{i}})\}_{i=1}^{n}$ and a sequence of word representations $\mathbf{e} = \{F_{W}(w_{i})\}_{i=1}^{n}$. Given that $G$ is the downstream model that takes a sequence of vectors as inputs and outputs prediction scores (e.g., logits), then we define two losses as follows:
\begin{equation}
\label{eq3}
    \mathcal{L}_{P} = \ell(Z_{P},y), \,\, \mathcal{L}_{W} = \ell(Z_{W},y),
\end{equation}
where $Z_{P}=G(\mathbf{p})$, $Z_{W}=G(\mathbf{e})$, and $\ell$ denotes a loss function such as the cross-entropy loss. By introducing $\mathcal{L}_{P}$ and $\mathcal{L}_{W}$, the downstream model is expected to focus on the morphological information in $\mathbf{e}$ and phonemic information in $\mathbf{p}$. In addition, we would like the model to focus on the connections between representations in distinct modalities. Therefore, we introduce an additional loss computed using a mixed sequence $\mathbf{m}=\!\{F(w_{i})\}_{i=1}^{n}$ as inputs:
\begin{equation}
\label{eq4}
    \mathcal{L}_{M} = \ell(Z_{M},y),
\end{equation}
where $Z_{M}=G(\mathbf{m})$. The loss to train the model $G$ is the sum of losses in Eq. \ref{eq3} and \ref{eq4}:
\begin{equation}
\label{eq5}
    \mathcal{L} = \alpha_{1}\mathcal{L}_{P}+\alpha_{2}\mathcal{L}_{W}+\alpha_{3}\mathcal{L}_{M},
\end{equation}
where $\alpha_{1}$, $\alpha_{2}$, and $\alpha_{3}$ are tuned hyperparameters to balance between the three losses.

\subsubsection{Cross-modal Ensemble}
\label{subsubsec:ensemble}
As explained in the previous subsection, we use phoneme, word, and mixed representations as parallel inputs to the downstream model, encouraging it to focus on different types of information in the representations. Hence, the model can perform better during inference if it simultaneously accesses such diverse types of useful information. Based on this intuition, we borrow the idea of a model ensemble and adopt a weighted average of the prediction scores as the final prediction score:
\begin{equation}
\label{eq6}
    Z = \beta_{1} Z_{P} + \beta_{2} Z_{W} + \beta_{3} Z_{M},
\end{equation}
where $\beta_{1}$, $\beta_{2}$, and $\beta_{3}$ are fixed weights of prediction scores.

\section{Experimental Setup}

\subsection{Datasets and Evaluation}
\label{subsec:dataset}
To evaluate the performance of KOPL on downstream tasks, we select popular Korean datasets that contain a large portion of real-world OOV words: KOLD~\cite{kold_EMNLP2022}, KLUE~\cite{park2021klue}, and NSMC.\footnote{\url{https://github.com/e9t/nsmc}} We use KOLD, KLUE-TC, and NSMC for sentence classification and use KLUE-DP and KLUE-NER for sequence tagging. For the evaluation metrics, we report macro F1 on KOLD and KLUE-TC, accuracy on NSMC, LAS on KLUE-DP, and entity-level macro F1 on KLUE-NER, consistent with~\cite{park-etal-2018-subword,park2021klue}. 

To demonstrate the improvement of our model on the OOV problem, we consider subsets of each dataset as additional evaluation sets, consisting of samples that include OOV words. Specifically, a word is considered as an OOV word if it is not included in the vocabulary of SISG(jm)~\cite{park-etal-2018-subword} containing 1.13M words. Since words so identified as OOV words are rare or unseen in a wide range of real-world corpora including encyclopedias, novels, newspapers, and dialogue~\cite{kim-etal-2022-break}, they are likely to be OOV words encountered in real cases. In the rest of the paper, we call the subsets constructed in this way as OOV subsets of the datasets. The statistics of datasets are given in Table \ref{tab:statistics}.

\begin{table}
\begin{center}
\setlength\tabcolsep{4.2pt}
{\small
\begin{tabular}{llrrrrr}
\toprule
Dataset & Task & Train & Dev & Test & OOV & \# Class \\ \midrule
KOLD & Offensive Detection & 32.3k & 4k & 4k & 0.5k & 2 \\
KLUE-TC & Topic Classification & 45.7k & 9.1k & 9.1k & 2.4k & 7 \\
NSMC & Sentiment Analysis & 125k & 25k & 50k & 8.5k & 2 \\
KLUE-DP & Dependency Parsing & 10k & 2k & 2.5k & 0.5k & 38 \\
KLUE-NER & Named Entity Recognition & 21k & 5k & 5k & 1.8k & 12 \\ \bottomrule
\end{tabular}
}
\end{center}
\caption{Statistics of each dataset. ``OOV'' denotes the number of examples in the OOV subset of the test set.}
\label{tab:statistics}
\end{table}

\subsection{Baselines}
\label{subsec:baseline}
For a comprehensive and fair comparison, we consider two categories of baseline models, which have been proposed to generate embeddings for OOV words. The first category of models are the methods designed to address Korean OOV words including SISG(jm)~\cite{park-etal-2018-subword}, misK~\cite{kwon-etal-2021-handling}, and SISG(BTS)~\cite{kim-etal-2022-break}. The second category of models are the methods originally tailored to English OOV words including MIMICK~\cite{MIMICK_pinter17}, LOVE~\cite{LOVE_chen22}, and GRM~\cite{GRM-2023-graph}. We train the baselines in this category by applying Korean-specific input methods as described in Table \ref{tab:models}, enabling them to process OOV words in Korean datasets. Additionally, to verify that the effectiveness of our method mainly comes from using phoneme representations, rather than the ensemble strategy, we consider another baseline called Ensemble, an ensemble model where we employ three distinct checkpoints of LOVE pre-trained from different random seeds.

\begin{table}
\begin{center}
\setlength\tabcolsep{4.0pt}
{\small
\begin{tabular}{lccc}
\toprule
Model & Input & Example \\ \midrule
MIMICK~\cite{MIMICK_pinter17} & subcharacter & \{ㅁ,ㅏ,ㅅ,ㅇ,ㅣ,ㅆ,ㄷ,ㅏ\} \\
LOVE~\cite{LOVE_chen22} & subcharacter + morpheme & \{ㅁ,ㅏ,ㅅ,ㅇ,ㅣ,ㅆ,ㄷ,ㅏ\} $\cup$ \{맛있,\#\#다\} \\ 
GRM~\cite{GRM-2023-graph} & morpheme & \{맛있,\#\#다\} \\ \bottomrule
\end{tabular}
}
\end{center}
\caption{Korean-specific input methods of different baselines originally tailored to English OOV words, with the Korean word 맛있다\textsubscript{delicious} as an example.}
\label{tab:models}
\end{table}


\subsection{Implementation Detail}
\label{subsec:det_ours}
In this subsection, we describe details of how to implement KOPL. Specifically, to implement $F_{P}$ and $F_{W}$, we adopt a 2-layer self-attention model~\cite{Vaswani_NIPS17} whose hidden dimension equals 300. To construct inputs to $F_{P}$, we straightforwardly use a list of IPA letters of each word (e.g., \{m,\textipa{A},s,i,t,t',\textipa{A}\} for the word ``맛있다\textsubscript{delicious}''). In addition, $F_{W}$ takes a mixed input of subcharacters and morphemes, similar to the input method for the Korean version of LOVE~\cite{LOVE_chen22} described in Table \ref{tab:models}. 

For the linear interpolation described in Eq. \ref{eq1}, we set the mixup ratio $\lambda$ to 0.1. We use the pre-trained word embeddings from SISG(jm) \cite{park-etal-2018-subword} as target word embeddings (i.e., $\mathbf{v}_{w}$ in Section \ref{subsec:phonemevec}). We use a temperature $\tau=0.07$ for the contrastive loss in Eq. \ref{eq2}. In addition, we apply AdamW as the optimizer and train the model for 10 epochs. When fine-tuning the models on downstream tasks, we freeze the input vectors generated by our method, consistent with the prior work~\cite{LOVE_chen22}. For the architecture of downstream models, we employ 1-layer CNN for KOLD, KLUE-TC, and NSMC, BiLSTM for KLUE-DP, and BiLSTM with CRF~\cite{huang_2015} for KLUE-NER. We use $\alpha_{1}=\alpha_{2}=\alpha_{3}=1$ and $\beta_{1}=\beta_{2}=\beta_{3}=1/3$ during fine-tuning.

\section{Results and Analysis}

\subsection{Main Results}
We report the results on the OOV subsets and their original counterparts in Table \ref{tab:main-table-extrinsic}. We can observe that KOPL achieves the best performance on both types of datasets. Moreover, KOPL consistently attains higher performance gains than the ensemble baseline across all datasets. This indicates that using phoneme representations in combination with word representations is helpful in handling Korean OOV words, by capturing the correlation between Korean letters and phonemes.

On average, the Korean version of English OOV methods MIMICK, LOVE, and GRM fall behind or are at most on par with the state-of-the-art Korean OOV method SISG(BTS). These results suggest that handling Korean OOV words can be challenging even with the recent OOV methods that have originally focused on English. On the other hand, KOPL shows a significant performance improvement over SISG(BTS) by 2.6\% on the OOV subsets. This indicates KOPL successfully makes downstream models more robust to Korean OOV words by considering the linguistic property of Korean, i.e., the correlation between letters and phonemes.

\begin{table*}
\begin{center}
\setlength\tabcolsep{2.0pt}
{\small
\begin{tabular}{lcccccccccccc}
\toprule
\multirow{2}{*}{Model} & \multicolumn{2}{c}{\makecell[c]{KOLD \\ (F1 $\uparrow$)}} & \multicolumn{2}{c}{\makecell[c]{KLUE-TC \\ (F1 $\uparrow$)}} &  \multicolumn{2}{c}{\makecell[c]{NSMC \\ (Acc. $\uparrow)$}} & \multicolumn{2}{c}{\makecell[c]{KLUE-DP \\ (LAS $\uparrow$)}} & \multicolumn{2}{c}{\makecell[c]{KLUE-NER \\ (F1 $\uparrow$)}} & \multicolumn{2}{c}{Avg.} \\ \cmidrule(lr){2-3}\cmidrule(lr){4-5}\cmidrule(lr){6-7}\cmidrule(lr){8-9}\cmidrule(lr){10-11}\cmidrule(lr){12-13}
& OOV & Orig. & OOV & Orig. & OOV & Orig. & OOV & Orig. & OOV & Orig. & OOV & Orig. \\ \midrule

SISG(jm)~\cite{park-etal-2018-subword} & 70.6 & 76.5 & 77.9 & 79.2 & 78.4 & 83.7 & 78.1 & 81.7 & 76.8 & 80.8 & 76.0 & 80.1 \\
misK~\cite{kwon-etal-2021-handling} & 73.1 & 74.5 & 65.8 & 66.1 & 79.2 & 82.7 & 77.9 & 80.2 & 74.3 & 75.5 & 74.0 & 75.8 \\ 
SISG(BTS)~\cite{kim-etal-2022-break} & 71.4 & 75.4 & 76.3 & 78.0 & 78.9 & 84.0 & 78.5 & 82.1 & 81.9 & 83.8 & 77.4 & 80.7 \\ \midrule

MIMICK~\cite{MIMICK_pinter17} & 64.9 & 70.1 & 51.7 & 51.7 & 69.1 & 74.2 &  77.3 & 80.7 & 69.1 & 73.4 & 66.4 & 70.0 \\
LOVE~\cite{LOVE_chen22} & 72.2 & 75.1 & 73.9 & 75.8 & 78.3 & 83.1 & 81.1 & 83.1 & 82.7 & 83.6 & 77.6 & 80.2 \\
GRM~\cite{GRM-2023-graph} & 69.8 & 74.5 & 72.0 & 75.5 & 74.9 & 81.6 & 75.9 & 78.7 & 76.0 & 78.9 & 73.7 & 77.8 \\ 
Ensemble & 72.8 & 75.5 & 74.7 & 76.7 & 78.5 & 83.3 & 81.5 & 83.3 & 83.0 & 83.7 & 78.1 & 80.5 \\ \midrule
KOPL (Ours) & \textbf{76.5} & \textbf{77.0} & \textbf{76.8} & \textbf{78.4} & \textbf{79.8} & \textbf{84.0} & \textbf{82.9} & \textbf{84.2} & \textbf{83.8} & \textbf{84.4} & \textbf{80.0} & \textbf{81.6} \\ \bottomrule
\end{tabular}
}
\end{center}
\caption{Performances on the downstream tasks. ``OOV'' and ``Orig.'' denote the OOV subset of the dataset and its original counterpart, respectively.}
\label{tab:main-table-extrinsic}
\end{table*}

\subsection{Ablation Study}
\label{subsec:ablation}
As we fine-tune the model in our approach, we use three types of inputs: phoneme, word, and mixed representations. To verify the effectiveness of using each type of input representation, we conduct an ablation study\footnote{To ablate the input representations of a specific type, we remove the loss and prediction score of that type from Eq. \ref{eq5} and Eq. \ref{eq6}, respectively (e.g., $\mathcal{L}_{P}$ and $Z_{P}$ for phoneme).} and report the results in Table \ref{tab:ablation}. Compared to using word representations only (row 2), utilizing phoneme representations (row 4) is helpful for improving performances on both original datasets and their OOV subsets. In addition, omitting phoneme representations (row 6) leads to significant performance degradation on both types of datasets, when compared to the full model (row 7). This means that learning information contained in phoneme representations is crucial for making the model robust to Korean OOV words.

When we do not use word representations (row 5), we observe a performance decrease on both original datasets and their OOV subsets, compared to the full model (row 7). This indicates that we still need word representations to achieve better performance. Moreover, comparing rows 5-7, the model achieves the lowest performance on both types of datasets when we exclude mixed representations (row 4). This suggests that capturing the connection between phoneme and word representations is important for boosting the performance, which can especially be the case for Korean that uses phonemic script \cite{sampson1985writing}, where letters exhibit a strong correlation with phonemes. 

Overall, as shown in row 7, using all three types of representations achieves the best results on both the original datasets and their OOV subsets. This indicates that each type of representation contains distinct information and that utilizing them simultaneously improves the model robustness, confirming our intuition described in Section \ref{subsubsec:ensemble}.

\begin{table*}
\begin{center}
\setlength\tabcolsep{2.6pt}
{\small
\begin{tabular}{c|ccc|cc|cc|cc|cc|cc}
\toprule
 & \multicolumn{3}{c|}{\makebox[0pt]{Input Representation}} & \multicolumn{2}{c|}{\makecell[c]{KOLD \\ (F1 $\uparrow$)}} & \multicolumn{2}{c|}{\makecell[c]{KLUE-TC \\ (F1 $\uparrow$)}} &  \multicolumn{2}{c|}{\makecell[c]{NSMC \\ (Acc. $\uparrow$)}} & \multicolumn{2}{c|}{\makecell[c]{KLUE-DP \\ (LAS $\uparrow$)}} & \multicolumn{2}{c}{\makecell[c]{KLUE-NER \\ (F1 $\uparrow$)}} \\ 
 \midrule
& Phoneme & Word & Mixed & OOV & Orig. & OOV & Orig. & OOV & Orig. & OOV & Orig. & OOV & Orig. \\ \midrule
1 & \ding{51} & & & 68.5 & 70.8 & 29.0 & 27.6 & 66.7 & 69.4 & 75.0 & 78.5 & 54.2 & 60.1 \\ 
2 & & \ding{51} & & 72.2 & 75.1 & 73.9 & 75.8 & 78.3 & 83.1 & 81.1 & 83.1 & 82.7 & 83.6 \\ 
3 & & & \ding{51} & 73.7 & 75.9 & 74.6 & 76.5 & 78.5 & 83.4 & 81.3 & 83.2 & 82.9 & 83.7 \\ 
4 & \ding{51} & \ding{51} & & 74.2 & 76.0 & 74.0 & 76.0 & 78.2 & 83.1 & 81.1 & 83.2 & 82.9 & 83.7 \\
5 & \ding{51} & & \ding{51} & 75.5 & 76.5 & 75.3 & 76.9 & 79.1 & 83.4 & 82.4 & 83.7 & 83.4 & 84.0 \\ 
6 & & \ding{51} & \ding{51} & 74.9 & 76.2 & 74.6 & 76.6 & 78.5 & 83.4 & 81.5 & 83.3 & 83.2 & 83.8 \\ 
7 & \ding{51} & \ding{51} & \ding{51} & \textbf{76.5} & \textbf{77.0} & \textbf{76.8} & \textbf{78.4} & \textbf{79.8} & \textbf{84.0} & \textbf{82.9} & \textbf{84.2} & \textbf{83.8} & \textbf{84.4} \\  \bottomrule
\end{tabular}
}
\end{center}
\caption{Performances on the downstream tasks with different types of input representations used to fine-tune downstream models. ``OOV'' and ``Orig.'' denote the OOV subset of the dataset and its original counterpart, respectively.}

\label{tab:ablation}
\end{table*}

\subsection{Model Adaptability}
\label{subsec:adapt}
To demonstrate the adaptability of our model, we combine KOPL with three existing Korean embedding models: SISG(jm)~\cite{park-etal-2018-subword}, SISG(BTS)~\cite{kim-etal-2022-break}, and BERT(Morph.)~\cite{Park_AACL2020}. It is worth noting that the first two baselines are Korean variants of FastText~\cite{fastText_bojanowski17} and produce static word embeddings, whereas the last one dynamically generates embeddings based on context. 

To combine KOPL with these baselines during fine-tuning on downstream tasks, we redefine input representation sequences $\mathbf{p}$, $\mathbf{e}$, and $\mathbf{m}$ originally defined in Section \ref{subsec:utilphoneme}, using the representations obtained from both the baseline and KOPL. The specific way of redefining input representation sequences depends on whether the baseline uses static or contextual word embeddings. In particular, for the static word embedding model, we use the word representations of the baseline and the phoneme representations of KOPL during fine-tuning. For the contextual word embedding model, we follow the approach of~\cite{LOVE_chen22}, where the OOV word is first identified and then the sub-word embeddings of that OOV word are replaced by an embedding generated by the OOV method.

As seen in Table \ref{tab:main-table-extrinsic2}, our model brings consistent performance gains on all datasets when combined with each baseline. This demonstrates that KOPL can be employed in a plug-and-play manner to make existing Korean embedding models more robust to OOV words. In particular, the improvement is attributed to modeling the strong correlation between Korean letters and phonemes, which have been overlooked by the prior Korean word embedding works~\cite{kim-etal-2022-break,park-etal-2018-subword}.

\begin{table*}[h]
\centering
{\small
\setlength{\tabcolsep}{1.0pt}         
\begin{tabular}{lcccccccccccc}
\toprule
\multirow{2}{*}{Model} & \multicolumn{2}{c}{\makecell[c]{KOLD \\ (F1 $\uparrow$)}} & \multicolumn{2}{c}{\makecell[c]{KLUE-TC \\ (F1 $\uparrow$)}} &  \multicolumn{2}{c}{\makecell[c]{NSMC \\ (Acc. $\uparrow$)}} & \multicolumn{2}{c}{\makecell[c]{KLUE-DP \\ (LAS $\uparrow$)}} & \multicolumn{2}{c}{\makecell[c]{KLUE-NER \\ (F1 $\uparrow$)}} & \multicolumn{2}{c}{Avg.} \\ \cmidrule(lr){2-3}\cmidrule(lr){4-5}\cmidrule(lr){6-7}\cmidrule(lr){8-9}\cmidrule(lr){10-11}\cmidrule(lr){12-13}
& OOV & Orig. & OOV & Orig. & OOV & Orig. & OOV & Orig. & OOV & Orig. & OOV & Orig. \\ \midrule

SISG(jm)~\cite{park-etal-2018-subword} & 70.6 & 76.5 & 76.1 & 77.9 & 78.4 & 83.7& 78.1 & 81.7 & 76.8 & 80.8 & 76.0 & 80.1 \\
SISG(jm) + KOPL & \textbf{75.6} & \textbf{77.4} & \textbf{79.3} & \textbf{80.0} & \textbf{79.7} & \textbf{83.9} & \textbf{80.9} & \textbf{83.2} & \textbf{79.1} & \textbf{82.2} & \textbf{78.9} & \textbf{81.3} \\ \midrule
SISG(BTS)~\cite{kim-etal-2022-break} & 71.4 & 75.4 & 76.3 & 78.0 & 78.9 & 84.0 & 78.5 & 82.1 & 81.9 & 83.8 & 77.4 & 80.7 \\
SISG(BTS) + KOPL & \textbf{73.3} & \textbf{76.2} & \textbf{79.5} & \textbf{80.1} & \textbf{81.0} & \textbf{84.3} & \textbf{81.1} & \textbf{83.0} & \textbf{83.2} & \textbf{84.4} & \textbf{79.6} & \textbf{81.6} \\ \midrule
BERT(Morph.)~\cite{Park_AACL2020} & 79.4 & 79.6 & 82.9 & 84.2 & 87.5 & 89.5 & 85.1 & 86.6 & 84.5 & 86.1 & 83.8 & 85.2 \\
BERT(Morph.) + KOPL & \textbf{81.6} & \textbf{80.3} & \textbf{83.8} & \textbf{84.5} & \textbf{88.1} & \textbf{89.7} & \textbf{86.0} & \textbf{87.0} & \textbf{86.0} & \textbf{86.4} & \textbf{85.1} & \textbf{85.6} \\ \bottomrule
\end{tabular}
}
\caption{The results of extending existing Korean embedding models to the downstream datasets. ``OOV'' and ``Orig.'' denote the OOV subset of the dataset and its original counterpart, respectively.}
\label{tab:main-table-extrinsic2}
\end{table*}

\section{Related Work}
\subsection{Representation Learning for Out-of-Vocabulary}
A line of research has utilized the internal structure of words to generate representations for OOV words. Specifically, character-level subword information is typically used to obtain OOV word representations~\cite{fastText_bojanowski17,GRM-2023-graph,ustun-etal-2018-characters,wieting-etal-2016-charagram,zhu-etal-2019-systematic}. Several studies have further combined contextual information to derive more accurate representations~\cite{hu-etal-2019-shot,schick-schutze-2019-attentive,Schick_AAAI2019}. However, these approaches are limited in that they require considerable computational resources due to the need for pre-training from scratch to construct their word embeddings. As an alternative method,~\cite{MIMICK_pinter17} has proposed a character-level recurrent neural network that mimics pre-trained word embeddings.~\cite{LOVE_chen22} has proposed a contrastive learning framework that predicts the embeddings of original words given corrupted words. This approach is highly efficient in training, since it does not require pre-training on massive corpora. These two lines of research are rooted in the linguistic properties of English, such as English morphemes represented as character n-grams~\cite{fastText_bojanowski17} or English word formation rules~\cite{LOVE_chen22,GRM-2023-graph}.

In contrast, the Korean letter, \textit{Hangeul}, has a systematic internal structure correlated with the feature of phonemes~\cite{sampson1985writing}, fundamentally different from English. The consonants and vowels of Korean (i.e., \textit{jamo}) are composed of basic units of \textit{Hangeul}, which are related to phonemic features \cite{kim-etal-2022-break}. Considering linguistic properties of Korean,~\cite{park-etal-2018-subword} and~\cite{kim-etal-2022-break} have proposed subword representations using \textit{jamo} and basic units, respectively, showing effectiveness in handling Korean OOV words. 

\subsection{Phoneme Information for NLP}
Recently, there have been attempts to incorporate phoneme information into NLP models. For example, RoCBert~\cite{rocbert_acl2022} and PhonemeBERT~\cite{phonemeBERT} have modeled texts and phoneme sequences to make the models robust against perturbations in textual and speech input, respectively. Unlike our approach, these methods require pre-training dedicated transformers on large-scale data. Other studies~\cite{Li_EMNLP2022,Liang_ACL2023,pilan-volodina-2018-exploring,Zhang_ACL2021} have used textual input and phoneme information to address the spelling correction task, a possible approach to handle misspellings that may be identified as OOV words. However, this approach represents a distinct direction of ours, as our approach studies how to learn representations for OOV words without correcting their spellings.

\section{Conclusions}
We have presented a novel OOV framework, KOPL, to derive Korean OOV word representations based on the linguistic property of Korean as the phonemic script. KOPL learns representations of phonemes and words, thereby combine both types of representations to connect with the word meaning. KOPL performs better than other OOV methods across various Korean NLP tasks involving real-word OOV words. In addition, we have designed KOPL in a plug-and-play manner to easily combine it with existing models. Future work includes extending KOPL to other languages and exploring its integration with large language models to handle real-world OOV words.

\subsubsection{\ackname} This work was supported by the National Research Foundation of Korea (NRF) grant funded by the Korea government (MSIT) (No. RS-2024-00415812) and Institute of Information \& communications Technology Planning \& Evaluation (IITP) grant funded by the Korea government (MSIT) (No. RS-2024-00439328, Karma: Towards Knowledge Augmentation for Complex Reasoning (SW Starlab), No. RS-2024-00457882, AI Research Hub Project, and No. RS2019-II190079, Artificial Intelligence Graduate School Program (Korea University)). The work of the first two authors was also supported by Basic Science Research Program through the National Research Foundation of Korea (NRF) funded by the Ministry of Education (No. RS-2023-00271662 and RS-2024-00414396).

%
%
%
\bibliographystyle{splncs04}
\bibliography{paper-698}

\begin{thebibliography}{10}
\providecommand{\url}[1]{\texttt{#1}}
\providecommand{\urlprefix}{URL }
\providecommand{\doi}[1]{https://doi.org/#1}

\bibitem{IPA_1999}
Association, I.P.: Handbook of the International Phonetic Association: A guide to the use of the International Phonetic Alphabet. Cambridge University Press (1999)

\bibitem{fastText_bojanowski17}
Bojanowski, P., Grave, E., Joulin, A., Mikolov, T.: Enriching word vectors with subword information. Trans. Assoc. Comput. Linguistics  \textbf{5},  135--146 (2017)

\bibitem{psychology_carroll1986}
Carroll, D.W.: Psychology of language. Thomson Brooks/Cole Publishing Co (1986)

\bibitem{LOVE_chen22}
Chen, L., Varoquaux, G., Suchanek, F.M.: Imputing out-of-vocabulary embeddings with {LOVE} makes languagemodels robust with little cost. In: {ACL} 2022. pp. 3488--3504 (2022)

\bibitem{hu-etal-2019-shot}
Hu, Z., Chen, T., Chang, K., Sun, Y.: Few-shot representation learning for out-of-vocabulary words. In: {ACL} 2019. pp. 4102--4112 (2019)

\bibitem{huang_2015}
Huang, Z., Xu, W., Yu, K.: Bidirectional {LSTM-CRF} models for sequence tagging. CoRR  \textbf{abs/1508.01991} (2015)

\bibitem{kold_EMNLP2022}
Jeong, Y., Oh, J., Lee, J., Ahn, J., Moon, J., Park, S., Oh, A.: {KOLD:} korean offensive language dataset. In: {EMNLP} 2022. pp. 10818--10833 (2022)

\bibitem{Jin_AAAI2020}
Jin, D., Jin, Z., Zhou, J.T., Szolovits, P.: Is {BERT} really robust? {A} strong baseline for natural language attack on text classification and entailment. In: {AAAI} 2020. pp. 8018--8025 (2020)

\bibitem{kim-etal-2022-break}
Kim, N., Park, J., Choi, J., Jeon, E., Kang, Y., Lee, S.: Break it down into {BTS:} basic, tiniest subword units for korean. In: {EMNLP} 2022. pp. 7007--7024 (2022)

\bibitem{kim-etal-2024-kombo}
Kim, S., Park, J., Kim, Y., Lee, S.: {KOMBO}: {K}orean character representations based on the combination rules of subcharacters. In: Findings of the Association for Computational Linguistics ACL 2024. pp. 5102--5119 (2024)

\bibitem{kwon-etal-2021-handling}
Kwon, O., Kim, D., Lee, S., Choi, J., Lee, S.: Handling out-of-vocabulary problem in hangeul word embeddings. In: {EACL} 2021. pp. 3213--3221 (2021)

\bibitem{lee-etal-2024-length}
Lee, J., Moon, H., Lee, S., Park, C., Eo, S., Ko, H., Seo, J., Lee, S., Lim, H.: Length-aware byte pair encoding for mitigating over-segmentation in {K}orean machine translation. In: Findings of the Association for Computational Linguistics ACL 2024. pp. 2287--2303 (2024)

\bibitem{Li_EMNLP2022}
Li, J., Wang, Q., Mao, Z., Guo, J., Yang, Y., Zhang, Y.: Improving chinese spelling check by character pronunciation prediction: The effects of adaptivity and granularity. In: {EMNLP} 2022. pp. 4275--4286 (2022)

\bibitem{Liang_IJCAI2018}
Liang, B., Li, H., Su, M., Bian, P., Li, X., Shi, W.: Deep text classification can be fooled. In: {IJCAI} 2018. pp. 4208--4215 (2018)

\bibitem{Liang_ACL2023}
Liang, Z., Quan, X., Wang, Q.: Disentangled phonetic representation for chinese spelling correction. In: {ACL} 2023. pp. 13509--13521 (2023)

\bibitem{GRM-2023-graph}
Liang, Z., Lu, Y., Chen, H., Rao, Y.: Graph-based relation mining for context-free out-of-vocabulary word embedding learning. In: {ACL} 2023. pp. 14133--14149 (2023)

\bibitem{Park_AACL2020}
Park, K., Lee, J., Jang, S., Jung, D.: An empirical study of tokenization strategies for various korean {NLP} tasks. In: {AACL/IJCNLP} 2020. pp. 133--142 (2020)

\bibitem{park-etal-2018-subword}
Park, S., Byun, J., Baek, S., Cho, Y., Oh, A.: Subword-level word vector representations for korean. In: {ACL} 2018. pp. 2429--2438 (2018)

\bibitem{park2021klue}
Park, S., Moon, J., Kim, S., Cho, W., Han, J., Park, J., Song, C., Kim, J., Song, Y., Oh, T.H., Lee, J., Oh, J., Lyu, S., Jeong, Y., Lee, I., Seo, S., Lee, D., Kim, H., Lee, M., Jang, S., Do, S., Kim, S., Lim, K., Lee, J., Park, K., Shin, J., Kim, S., Park, E.L., Oh, A., Ha, J., Cho, K.: {KLUE:} korean language understanding evaluation. In: NeurIPS 2021 (2021)

\bibitem{pilan-volodina-2018-exploring}
Pil{\'a}n, I., Volodina, E.: Exploring word embeddings and phonological similarity for the unsupervised correction of language learner errors. In: Proceedings of the Second Joint {SIGHUM} Workshop on Computational Linguistics for Cultural Heritage, Social Sciences, Humanities and Literature. pp. 119--128 (2018)

\bibitem{MIMICK_pinter17}
Pinter, Y., Guthrie, R., Eisenstein, J.: Mimicking word embeddings using subword rnns. In: {EMNLP} 2017. pp. 102--112 (2017)

\bibitem{sampson1985writing}
Sampson, G.: Writing systems. London, UK: H utchinson  (1985)

\bibitem{KVQFH_sasaki19}
Sasaki, S., Suzuki, J., Inui, K.: Subword-based compact reconstruction of word embeddings. In: {NAACL-HLT} 2019. pp. 3498--3508 (2019)

\bibitem{schick-schutze-2019-attentive}
Schick, T., Sch{\"{u}}tze, H.: Attentive mimicking: Better word embeddings by attending to informative contexts. In: {NAACL-HLT} 2019. pp. 489--494 (2019)

\bibitem{Schick_AAAI2019}
Schick, T., Sch{\"{u}}tze, H.: Learning semantic representations for novel words: Leveraging both form and context. In: {AAAI} 2019. pp. 6965--6973 (2019)

\bibitem{seo-etal-2023-chef}
Seo, J., Moon, H., Lee, J., Eo, S., Park, C., Lim, H.: {CHEF} in the language kitchen: A generative data augmentation leveraging {K}orean morpheme ingredients. In: {EMNLP} 2023. pp. 6014--6029 (2023)

\bibitem{sofroniev-coltekin-2018-phonetic}
Sofroniev, P., {\c{C}}{\"o}ltekin, {\c{C}}.: Phonetic vector representations for sound sequence alignment. In: Proceedings of the Fifteenth Workshop on Computational Research in Phonetics, Phonology, and Morphology. Brussels, Belgium (2018)

\bibitem{rocbert_acl2022}
Su, H., Shi, W., Shen, X., Zhou, X., Ji, T., Fang, J., Zhou, J.: Rocbert: Robust chinese bert with multimodal contrastive pretraining. In: {ACL} 2022. pp. 921--931 (2022)

\bibitem{Sun_Corr2020}
Sun, L., Hashimoto, K., Yin, W., Asai, A., Li, J., Yu, P.S., Xiong, C.: Adv-bert: {BERT} is not robust on misspellings! generating nature adversarial samples on {BERT}. arXiv preprint arXiv:2003.04985  (2020)

\bibitem{phonemeBERT}
Sundararaman, M.N., Kumar, A., Vepa, J.: Phonemebert: Joint language modelling of phoneme sequence and {ASR} transcript. In: Interspeech 2021. pp. 3236--3240 (2021)

\bibitem{ustun-etal-2018-characters}
{\"{U}}st{\"{u}}n, A., Kurfali, M., Can, B.: Characters or morphemes: How to represent words? In: Proceedings of The Third Workshop on Representation Learning for NLP, Rep4NLP@ACL 2018. pp. 144--153 (2018)

\bibitem{Vaswani_NIPS17}
Vaswani, A., Shazeer, N., Parmar, N., Uszkoreit, J., Jones, L., Gomez, A.N., Kaiser, L., Polosukhin, I.: Attention is all you need. In: {NIPS} 2017. pp. 5998--6008 (2017)

\bibitem{wieting-etal-2016-charagram}
Wieting, J., Bansal, M., Gimpel, K., Livescu, K.: Charagram: Embedding words and sentences via character n-grams. In: {EMNLP} 2016. pp. 1504--1515 (2016)

\bibitem{Zhang_ACL2021}
Zhang, R., Pang, C., Zhang, C., Wang, S., He, Z., Sun, Y., Wu, H., Wang, H.: Correcting chinese spelling errors with phonetic pre-training. In: Findings of the Association for Computational Linguistics: {ACL/IJCNLP} 2021. pp. 2250--2261 (2021)

\bibitem{BOS_zhao18}
Zhao, J., Mudgal, S., Liang, Y.: Generalizing word embeddings using bag of subwords. In: {EMNLP} 2018. pp. 601--606 (2018)

\bibitem{zhu2024taste}
Zhu, J., Yang, C., Samir, F., Islam, J.: The taste of {IPA:} towards open-vocabulary keyword spotting and forced alignment in any language. In: {NAACL} 2024. pp. 750--772 (2024)

\bibitem{zhu-etal-2019-systematic}
Zhu, Y., Vulic, I., Korhonen, A.: A systematic study of leveraging subword information for learning word representations. In: {NAACL-HLT} 2019. pp. 912--932 (2019)

\end{thebibliography}

\end{document}